\documentclass[letterpaper]{article} 
\usepackage{aaai23}  
\usepackage{times}  
\usepackage{helvet}  
\usepackage{courier}  
\usepackage[hyphens]{url}  
\usepackage{graphicx} 
\usepackage{natbib}  
\usepackage{caption} 
\usepackage{algorithm}
\usepackage{algorithmic}
\usepackage{multirow}
\usepackage{makecell}
\usepackage{amsmath,amssymb}
\usepackage{booktabs}
\usepackage{dsfont}
 
\usepackage{newfloat}
\usepackage{listings}
\DeclareCaptionStyle{ruled}{labelfont=normalfont,labelsep=colon,strut=off} 
\floatstyle{ruled}
\newfloat{listing}{tb}{lst}{}
\floatname{listing}{Listing}
\title{Self-Emphasizing Network for Continuous Sign Language Recognition }
\author{
    Lianyu Hu, Liqing Gao, Zekang liu, Wei Feng\thanks{Corresponding author}
}
\affiliations{
  Tianjin University, Tianjin 300350, China\\
  hly2021,lqgao,lzk100953@tju.edu.cn,wfeng@ieee.org\\
}

\begin{document}

\maketitle

\begin{abstract}
Hand and face play an important role in expressing sign language. Their features are usually especially leveraged to improve system performance. However, to effectively extract visual representations and capture trajectories for hands and face, previous methods always come at high computations with increased training complexity. They usually employ extra heavy pose-estimation networks to locate human body keypoints or rely on additional pre-extracted heatmaps for supervision. To relieve this problem, we propose a self-emphasizing network (SEN) to emphasize informative spatial regions in a self-motivated way, with few extra computations and without additional expensive supervision. Specifically, SEN first employs a lightweight subnetwork to incorporate local spatial-temporal features to identify informative regions, and then dynamically augment original features via attention maps. It's also observed that not all frames contribute equally to recognition. We present a temporal self-emphasizing module to adaptively emphasize those discriminative frames and suppress redundant ones. A comprehensive comparison with previous methods equipped with hand and face features demonstrates the superiority of our method, even though they always require huge computations and rely on expensive extra supervision. Remarkably, with few extra computations, SEN achieves new state-of-the-art accuracy on four large-scale datasets, PHOENIX14, PHOENIX14-T, CSL-Daily, and CSL. Visualizations verify the effects of SEN on emphasizing informative spatial and temporal features. Code is available at \url{https://github.com/hulianyuyy/SEN_CSLR}
\end{abstract}

\section{Introduction}
Sign language is one of the most commonly-used communication tools for the deaf community in their daily life. It mainly conveys information by both manual components (hand/arm gestures), and non-manual components (facial expressions, head movements, and body postures)~\cite{dreuw2007speech,ong2005automatic}. However, mastering this language is rather difficult and time-consuming for the hearing people, thus hindering direct communications between two groups. To relieve this problem, isolated sign language recognition tries to classify a video segment into an independent gloss\footnote{Gloss is the atomic lexical unit to annotate sign languages.}. Continuous sign language recognition (CSLR) progresses by sequentially translating image streams into a series of glosses to express a complete sentence, more prospective towards bridging the communication gap. 


\begin{figure}[t]
    \centering
    \includegraphics[width=\linewidth]{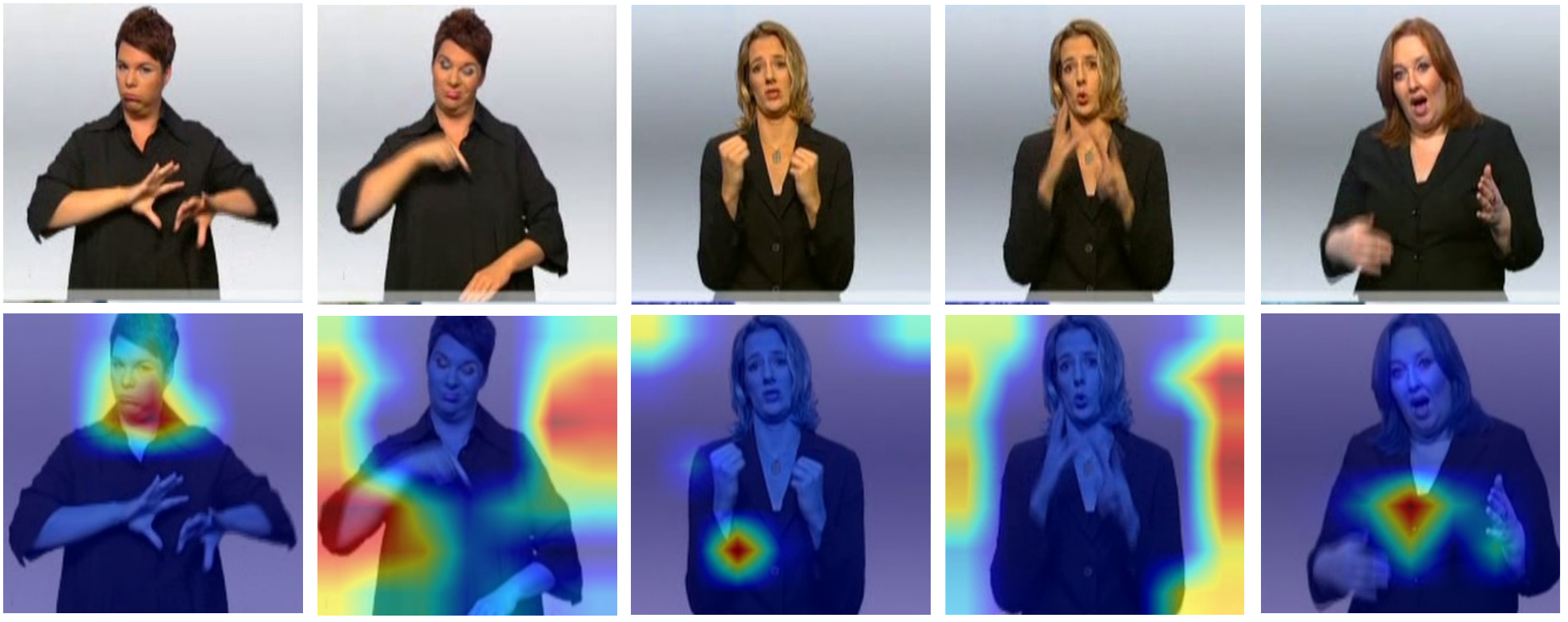}
    \caption{Visualization of class activation maps with Grad-CAM~\cite{selvaraju2017grad} for VAC~\cite{Min_2021_ICCV} (baseline). Top: Original frames. Bottom: activation maps. It's observed that without extra supervision, it fails to locate discriminative face and hand regions precisely. }
    \label{fig1}
  \end{figure}

In sign language, the left hand, right hand, and face play the most important role in expressing glosses. Mostly, they convey the information through horizontal/vertical hand movements, finger activities, and static gestures, assisted with facial expressions and mouth shapes to holistically deliver messages~\cite{dreuw2007speech,ong2005automatic}. As a result, hand and face, are always especially leveraged and incorporated in sign language systems. In isolated sign language recognition, early methods~\cite{freeman1995orientation,sun2013discriminative} leveraged hand-crafted features to describe the gestures and motion of both hands. Recent methods either choose to build a pure pose-based system~\cite{tunga2021pose,hu2021signbert} based on detected keypoints for both hands and face, or construct appearance-based systems~\cite{hu2021hand,boukhayma20193d} with cropped patches for hands and face as collaborative inputs. In CSLR, CNN-LSTM-HMM~\cite{koller2019weakly} builds a multi-stream (hands and face) Hidden-Markov-Model (HMM) to integrate multiple visual inputs to boost recognition accuracy. STMC~\cite{zhou2020spatial} explicitly inserts a pose-estimation network and uses the detected regions (hand and face) as multiple cues to perform recognition. More recently, C$^2$SLR~\cite{zuo2022c2slr} leverages the pre-extracted pose keypoints heatmaps as additional supervision to guide models to focus on hand and face areas. 

Although it has been proven effective to incorporate hand and face features to improve recognition performance for sign language systems, previous methods usually come at huge computations with increased training complexity, and rely on additional pose estimation networks or extra expensive supervision (e.g., heatmaps). However, without these supervision signals, we find current methods~\cite{Min_2021_ICCV,hao2021self,cheng2020fully} in CSLR fail to precisely locate the hand and face regions (Fig.~\ref{fig1}), thus unable to effectively leverage these features. To more effectively excavate these key cues but avoid introducing huge computations or relying on expensive supervision, we propose a self-emphasizing network (SEN) to explicitly emphasize informative spatial regions in a self-motivated way. Specifically, SEN first employs a lightweight subnetwork to incorporate local spatial-temporal features to identify informative regions, and then dynamically emphasizes or suppresses input features via attention maps. 

It's also observed that not all frames contribute equally to recognition. For example, frames with hand/arm movements of the signer are usually more important than those transitional frames. We present a temporal self-emphasizing module to emphasize those discriminative frames and suppress redundant ones dynamically. Remarkably, SEN yields new state-of-the-art accuracy upon four large-scale CSLR datasets, especially outperforming previous methods equipped with hand and face features, even though they always come at huge computations and rely on expensive supervision. Visualizations verify the effects of SEN in emphasizing spatial and temporal features.

\section{Related Work}
\subsection{Continuous Sign Language Recognition}
Sign language recognition methods can be roughly categorized into isolated sign language recognition~\cite{tunga2021pose,hu2021signbert,hu2021hand} and continuous sign language recognition~\cite{pu2019iterative,cheng2020fully,cui2019deep,niu2020stochastic,Min_2021_ICCV} (CSLR), and we focus on the latter in this paper. CSLR tries to translate image frames into corresponding glosses in a weakly-supervised way: only sentence-level label is provided. Early methods in CSLR usually depend on hand-crafted features~\cite{gao2004chinese,freeman1995orientation} to provide visual information, especially body gestures, hands, and face, or rely on HMM-based systems~\cite{koller2016deepsign,han2009modelling,koller2017re,koller2015continuous} to perform temporal modeling and then translate sentences step by step. The HMM-based methods typically first employ a feature extractor to capture visual representations and then adopt an HMM to perform long-term temporal modeling. The recent success of convolutional neural networks (CNNs) and recurrent neural networks brings huge progress for CSLR. The widely-used CTC loss~\cite{graves2006connectionist} enables end-to-end training for recent methods by aligning target glosses with inputs. 

Especially, hands and face are paid close attention to by recent methods. For example, CNN-LSTM-HMM~\cite{koller2019weakly} employs a multi-stream HMM (including hands and face) to integrate multiple visual inputs to improve recognition accuracy. STMC~\cite{zhou2020spatial} utilizes a pose-estimation network to estimate human body keypoints and then sends cropped patches (including hands and face) for integration. More recently, C$^2$SLR~\cite{zuo2022c2slr} leverages the pre-extracted pose keypoints as supervision to guide the model. Despite high accuracy, they consume huge additional computations and training complexity.

Practically, recent methods~\cite{pu2019iterative,pu2020boosting,cheng2020fully,cui2019deep,niu2020stochastic,Min_2021_ICCV} usually first employ a feature extractor to capture frame-wise visual representations for each frame, and then adopt 1D CNN and BiLSTM to perform short-term and long-term temporal modeling, respectively. However, several methods~\cite{pu2019iterative,cui2019deep} found in such conditions the feature extractor is not well trained and propose the iterative training strategy to refine the feature extractor, but consume much more computations. More recent methods try to directly enhance the feature extractor by adding visual alignment losses~\cite{Min_2021_ICCV} or adopt pseudo label~\cite{cheng2020fully,hao2021self} for supervision. We propose the self-emphasizing network to emphasize informative spatial features, which can be viewed to enhance the feature extractor in a self-motivated way.

\subsection{Spatial Attention }
Spatial attention has been proven to be effective in many fields including image classification~\cite{cao2019gcnet,hu2018gather,woo2018cbam,hu2018squeeze}, scene segmentation~\cite{fu2019dual} and video classification~\cite{wang2018non}. SENet~\cite{hu2018squeeze}, CBAM~\cite{woo2018cbam}, SKNet~\cite{li2019selective} and ECA-Net~\cite{wang2020eca} devise lightweight channel attention modules for image classification. The widely used self-attention operator~\cite{wang2018non} employs dot-product feature similarities to build attention maps and aggregate long-term dependencies. However, the calculation complexity of the self-attention operator is quadratic to the incorporated pixels, incurring a heavy burden for video-based tasks~\cite{wang2018non}. Instead of feature similarities, our SEN employs a learnable subnetwork to aggregate local spatial-temporal representations and generates spatial attention maps for each frame, much more lightweight than self-attention operators. Some works also propose to leverage external supervision to guide the spatial attention module. For example, GALA~\cite{linsley2018learning} collects click maps from games to supervise the spatial attention for image classification. A relation-guided spatial attention module~\cite{li2020relation} is designed to explore the discriminative regions globally for Video-Based Person Re-Identification. MGAN~\cite{pang2019mask} introduces an attention network to emphasize visible pedestrian regions by modulating full body features. In contrast to external supervision, our self-emphasizing network strengthens informative spatial regions in a self-motivated way, thus greatly lowering required computations and training complexity.

\section{Method}
\subsection{Framework Overview}
\begin{figure}[t]
\centering
\includegraphics[width=\linewidth]{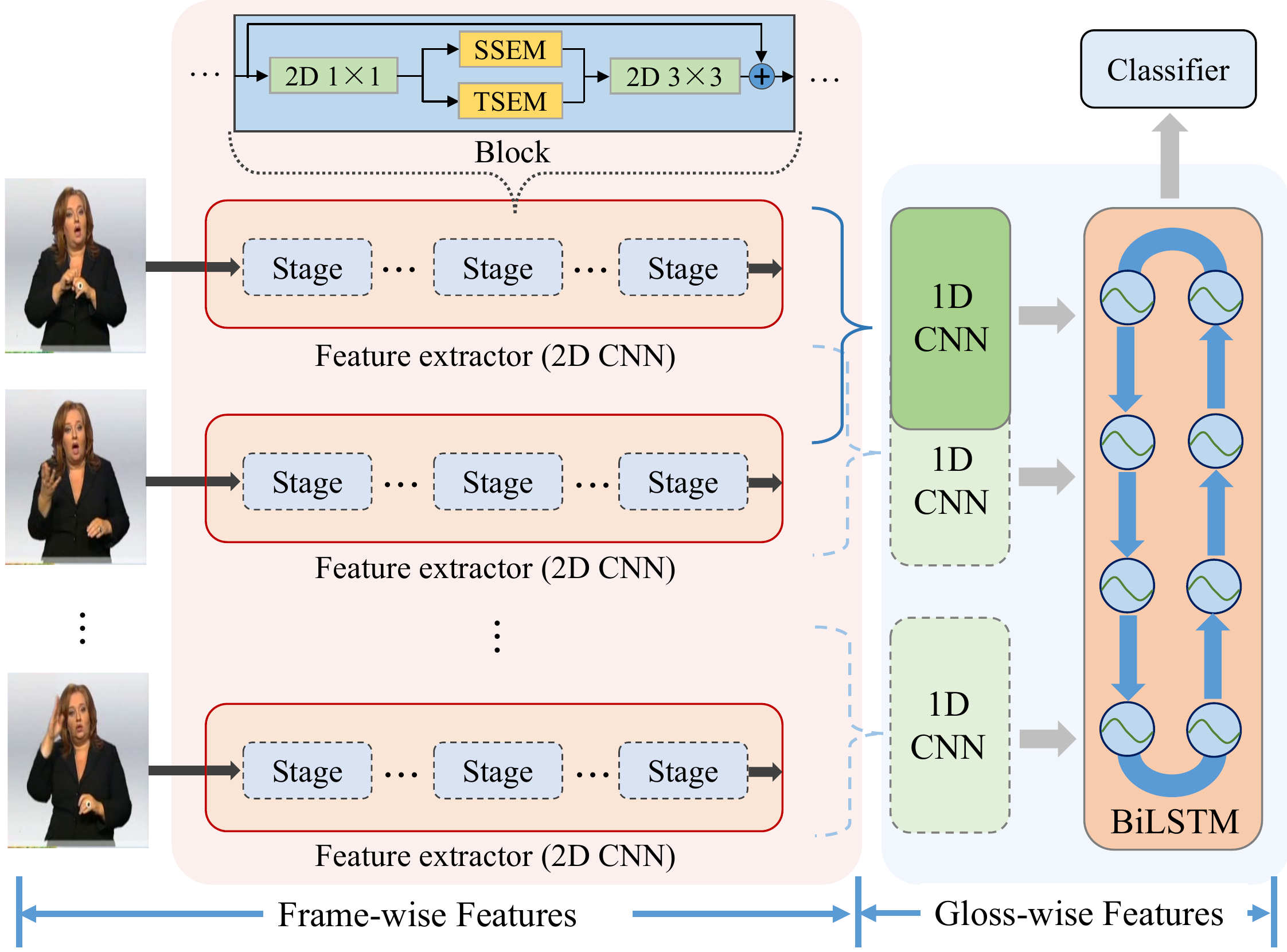}
\caption{A overview for our SEN. It first employs a feature extractor (2D CNN) to capture frame-wise features, and then adopts a 1D CNN and a BiLSTM to perform short-term and long-term temporal modeling, respectively, followed by a classifier to predict sentences. We place our proposed spatial self-emphasizing module (SSTM) and temporal self-emphasizing module (TSEM) into each block of the feature extractor to emphasize the spatial and temporal features, respectively.}
\label{fig2}
\end{figure}

As shown in fig.~\ref{fig2}, the backbone of CSLR models is consisted of a feature extractor (2D CNN\footnote{Here we only consider the feature extractor based on 2D CNN, because recent findings~\cite{adaloglou2021comprehensive,zuo2022c2slr} show 3D CNN can not provide as precise gloss boundaries as 2D CNN, and lead to lower accuracy. }), a 1D CNN, a BiLSTM, and a classifier (a fully connected layer) to perform prediction. Given a sign language video with $T$ input frames $x = \{x_{t}\}_{t=1}^T \in \mathcal{R}^{T \times 3\times H_0 \times W_0} $, a CSLR model aims to translate the input video into a series of glosses $y=\{ y_i\}_{i=1}^{N}$ to express a sentence, with $N$ denoting the length of the label sequence. Specifically, the feature extractor first processes input frames into frame-wise features $v = \{v_t\}_{t=1}^{T} \in \mathcal{R}^{T\times d}$. Then the 1D CNN and BiLSTM perform short-term and long-term temporal modeling based on these extracted visual representations, respectively. Finally, the classifier employs widely-used CTC loss to predict the probability of target gloss sequence $p(y|x)$.

To emphasize the informative spatial and temporal features for CSLR models, we present a spatial self-emphasizing module (SSEM) and a temporal self-emphasizing module (TSEM). Specifically, we incorporate them into the feature extractor to operate on each frame. Fig.~\ref{fig2} shows an example of a common feature extractor consisting of multiple stages with several blocks in each. We place the SSEM and TSEM in parallel before the $3\times 3$ spatial convolution in each block to emphasize informative spatial and temporal features, respectively. When designing the architecture, efficiency is our core consideration, to avoid heavy computational burdens like previous methods~\cite{zhou2020spatial,zuo2022c2slr} based on heavy pose-estimation networks or expensive heatmaps. We next introduce our SSEM and TSEM, respectively. 

\subsection{Spatial Self-Emphasizing Module (SSEM)}
From fig.~\ref{fig1}, we argue current CSLR models fail to effectively leverage the informative spatial features, e.g., hands and face. We try to enhance the capacity of the feature extractor of CSLR models to incorporate such discriminative features without affecting its original spatial modeling ability. Practically, our SSEM is designed to first leverage the closely correlated local spatial-temporal features to identify the informative regions for each frame, and then augment original representations in the form of attention maps.

\begin{figure}[t]
  \centering
  \includegraphics[width=0.8\linewidth]{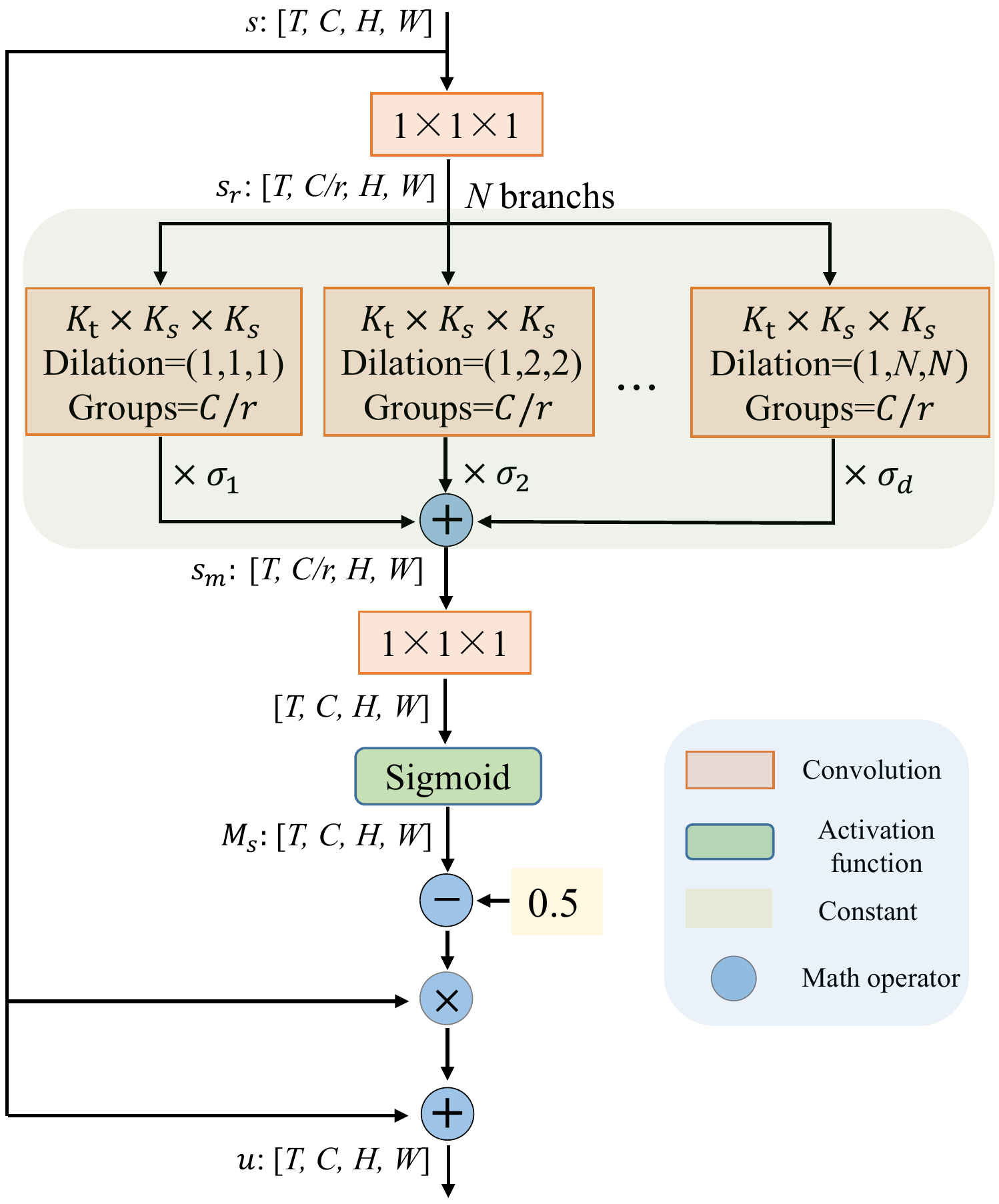}
  \caption{Illustration for our spatial self-emphasizing module (SSEM).} 
  \label{fig3}
  \end{figure}

As shown in fig.~\ref{fig3}, SSEM first projects the input features $s = \{s_t\}_{t=1}^T \in \mathcal{R}^{T \times C\times H \times W}$ into $s_r\in \mathcal{R}^{T \times C/r\times H \times W}$ to decrease the computational costs brought by SSEM, with $r$ the reduction factor as 16 by default.

The frame-wise features $s$ in the feature extractor are independently extracted for each frame by 2D convolutions, failing to incorporate local spatial-temporal features to distinguish the informative spatial regions. Besides, as the signer has to throw his/her arms and hands to express glosses, the informative regions in adjacent frames are always misaligned. Thus, we devise a multi-scale architecture to perceive spatial-temporal features in a large neighborhood to help identify informative regions. 

Instead of a large spatial-temporal convolution kernel, we employ $N$ parallel factorized branches with group-wise convolutions of progressive dilation rates to lower computations and increase the model capacity. As shown in fig.~\ref{fig3}, these $N$ branches own the same spatial-temporal kernel size $K_t\times K_s\times K_s$, with different spatial dilation rates $[1\cdots N]$. Features from different branches are multiplied with learnable factors $\{\sigma_{1}\dots \sigma_k\}$ to control the importance of different branches via gradient-based backward propagation, and are then added to mix information from different receptive fields. This multi-scale architecture is expressed as:

\begin{equation}
\label{e1}
s_m = \sum_{i=1}^{N}{\sigma_i \times {\rm Conv}_i(s_r)}
\end{equation}
where the group-wise convolution ${\rm Conv}_i$ at different levels captures spatial-temporal features from different receptive fields, with dilation rate $(1,i,i)$.

Especially, as the channels are downsized by $r$ times in SSEM and we employ group-wise convolutions with small spatial-temporal kernels to capture multi-scale features, the overall architecture is rather lightweight with few (\textless 0.1\%) extra computations compared to the original model, as demonstrated in our ablative experiments.

Next, $s_m$ is sent into a $1\times 1 \times 1$ convolution to project channels back into $C$, and then passed through a sigmoid activation function to generate attention maps $M_s\in \mathcal{R}^{T \times C\times H \times W}$ with values ranging between $[0, 1 ]$ as:
\begin{equation}
\label{e2}
M_s = {\rm Sigmoid}({\rm Conv}_{1\times 1\times 1}(s_m))
\end{equation}

Finally, the attention maps $M_s$ are used to emphasize informative spatial regions for input features. To avoid hurting original representations and degrading accuracy, we propose to emphasize input features via a residual way as:
\begin{equation}
\label{e3}
u = (M_s-0.5\times \mathds{1})\odot s+s
\end{equation}
where $\odot$ denotes element-wise multiplication and $u$ is the output. 

In specific, we first subtract $0.5\times \mathds{1}$ from the attention maps $M_s$, with $\mathds{1}\in \mathcal{R}^{T \times C\times H \times W}$ denoting an all-one matrix, to change the range of values in $M_s$ into $[-0.5, 0.5]$. Then we element-wisely multiply the resulting attention maps with input features $s$ to dynamically emphasize the informative regions and suppress unnecessary areas. Here, the values in $M_s$ larger than 0 would strengthen the corresponding inputs features, otherwise they would weaken the input features. Finally, we add the modulated features with input features $s$ to emphasize or suppress certain spatial features, but avoid hurting original representations.

\begin{figure}[t]
  \centering
  \includegraphics[width=0.7\linewidth]{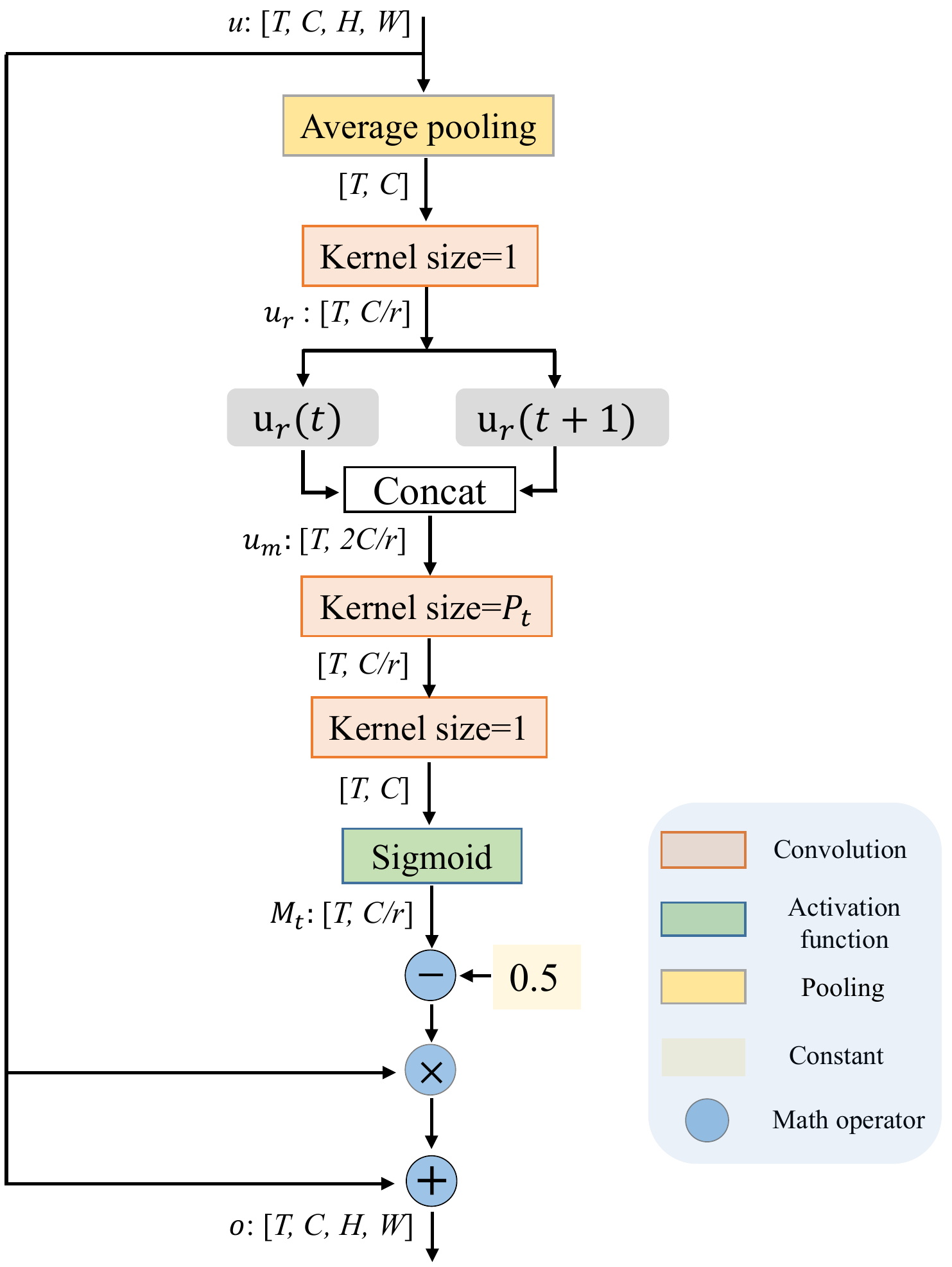}
  \caption{Illustration for our temporal self-emphasizing module (TSEM).} 
  \label{fig4}
  \end{figure}

\subsection{Temporal Self-Emphasizing Module}
We argue that not all frames in a video contribute equally to recognition, where some frames are more discriminative than others. For example, frames in which the signer moves his/her arms to express a sign are usually more important than those transitional frames or idle frames with meaningless contents. However, the feature extractor only employs 2D spatial convolutions to capture spatial features for each frame, equally treating frames without considering their temporal correlations. We propose a temporal self-emphasizing module (TSEM) to adaptively emphasize discriminative frames and suppress redundant ones. 

As shown in fig.~\ref{fig4}, input features $u \in \mathcal{R}^{T \times C\times H \times W}$ first undergo a global average pooling layer to eliminate the spatial dimension, i.e., $H$ and $W$. Then these features pass through a convolution with kernel size of 1 to reduce channels by $r$ times into $u_r \in \mathcal{R}^{T \times C/r}$ as:
\begin{equation}
  \label{e4}
  u_r = {\rm Conv}_{K=1}({\rm AvgPool}(u))
  \end{equation}
where $K$ denotes the kernel size. To better exploit local temporal movements to identify the discriminative frames, we leverage the temporal difference operator to incorporate motion information between adjacent frames to make decisions better. Specially, we calculate the difference between two adjacent frames for $u_r$ as approximate motion information, and then concatenate it with appearance features $u_r$ as : 
\begin{equation}
\label{e5}
u_m = {\rm Concat}([u_r, u_r(t+1)-u_r])
\end{equation}

Next, we send $u_m$ into a 1D temporal convolution with kernel size of $P_t$ to capture the short-term temporal information. As the size of $u_m$ is rather small, we here employ a normal temporal convolution instead of a multi-scale architecture. The features then undergo a convolution with kernel size of 1 to project channels back into $C$, and pass through a sigmoid activation function to generate attention maps $M_t \in \mathcal{R}^{T \times C}$ as:
\begin{equation}
  \label{e5}
  M_t = {\rm Sigmoid}({\rm Conv}_{K=1}(u_m))
  \end{equation}

Finally, we employ $M_t $ to emphasize the discriminative features for input $u$ in a residual way as :
\begin{equation}
  \label{e6}
  o = (M_t-0.5\times \mathds{1})\odot u+u
  \end{equation}
where $\odot$ denotes element-wise multiplication and $o$ is the output.


\section{Experiments}
\subsection{Experimental Setup}
\subsubsection{Datasets.} \textbf{PHOENIX14}~\cite{koller2015continuous} and \textbf{PHOENIX14-T}~\cite{camgoz2018neural} are both recorded from a German weather forecast broadcast before a clean background with a resolution of 210 $\times$ 260. They contain 6841/8247 sentences with a vocabulary of 1295/1085 signs, divided into 5672/7096 training samples, 540/519 development (Dev) samples and 629/642 testing (Test) samples.

\textbf{CSL-Daily}~\cite{zhou2021improving} is recorded indoor with 20654 sentences, divided into 18401 training samples, 1077 development (Dev) samples and 1176 testing (Test) samples. 

\textbf{CSL}~\cite{huang2018video} is collected in the laboratory environment by fifty signers with a vocabulary size of 178 with 100 sentences. It contains 25000 videos, divided into training and testing sets by a ratio of 8:2.

\subsubsection{Training details.} We adopt ResNet18~\cite{he2016deep} as the 2D CNN with ImageNet~\cite{deng2009imagenet} pretrained weights. We place SSEM and TSEM before the second convolution in each block. The 1D CNN consists of a sequence of \{K5, P2, K5, P2\} layers where $K$ and $P$ denotes a 1D convolutional layer and a pooling layer with kernel size of 5 and 2, respectively. We then adopt a two-layer BiLSTM with 1024 hidden states and a fully connected layer for prediction. We train our model for 80 epochs with initial learning rate 0.0001 decayed by 5 after 40 and 60 epochs. Adam optimizer is adopted with weight decay 0.001 and batch size 2. All frames are first resized to 256$\times$256 and then randomly cropped to 224$\times$224, with 50\% horizontal flip and $\pm$20\% random temporal scaling during training. During inference, a central 224$\times$224 crop is simply selected. We use VE and VA losses from VAC~\cite{Min_2021_ICCV} for extra supervision. 

\subsubsection{Evaluation Metric.} We use Word Error Rate (WER) as the evaluation metric, which is defined as the minimal summation of the \textbf{sub}stitution, \textbf{ins}ertion, and \textbf{del}etion operations to convert the predicted sentence to the reference sentence, as:
\begin{equation}
\label{e11}
\rm WER = \frac{ \#sub+\#ins+\#del}{\#reference}.
\end{equation}
Note that the \textbf{lower} WER, the \textbf{better} accuracy.

\begin{table}[t]   
  \centering
  \begin{tabular}{cccc}
  \hline
  Configurations & FLOPs & Dev(\%) & Test(\%)\\
  \hline
  - & 3.64G& 21.2 & 22.3\\
  $K_t$=9, $K_s$=3, $N$=\textbf{1} & +0.4M  & 20.5  & 22.0 \\
  $K_t$=9, $K_s$=3, $N$=\textbf{2} & +0.6M& 20.2 & 21.8 \\
  $K_t$=9, $K_s$=3, $N$=\textbf{3} & +0.8M & \textbf{19.9}  & \textbf{21.4} \\
  $K_t$=9, $K_s$=3, $N$=\textbf{4} & +1.0M & 20.2  & 21.7 \\
  $K_t$=\textbf{7}, $K_s$=3, $N$=3 & +0.7M & 20.2  & 21.6 \\
  $K_t$=\textbf{11}, $K_s$=3, $N$=3 & +1.0M & 20.3  & 21.8 \\
  $K_t$=9, $K_s$=\textbf{7}, $N$=1 & +2.9M & 20.5 & 22.0 \\
  \hline
  \end{tabular}
  \caption{Ablations for the multi-scale architecture of SSEM on the PHOENIX14 dataset.} 
  \label{tab1} 
  \end{table}

  \begin{table}[t]   
    \centering
    \begin{tabular}{ccc}
    \hline
    Configurations & Dev(\%) & Test(\%)\\
    \hline
    - & 21.2 & 22.3\\
    $M_s \odot s$  & 22.3  & 23.4 \\
    $M_s \odot s + s$ & 20.6 & 21.7 \\
    $(M_s -0.5\times \mathds{1}) \odot s$ & 20.2  & 21.5 \\
    $(M_s -0.5\times \mathds{1}) \odot s +s $ & \textbf{19.9}  & \textbf{21.4}\\
    \hline
    \end{tabular}
    \caption{Ablations for the implementations of SSEM to augment input features on the PHOENIX14 dataset.} 
    \label{tab3} 
    \end{table}

\begin{table}[t]   
  \centering
  \begin{tabular}{ccc}
  \hline
  Configurations & Dev(\%) & Test(\%)\\
  \hline
  - & 19.9 & 21.4 \\
  \hline
  $u_r$ & 19.8  & 21.2 \\
  ${\rm Concat}([u_r, u_r(t+1)-u_r])$ & \textbf{19.5}  & \textbf{21.0} \\
  \hline
  $P_t$ = 7 & 19.6  & 21.2 \\
  $P_t$ = 9 & \textbf{19.5} & \textbf{21.0} \\
  $P_t$ = 11 & 19.7  & 21.3 \\
  \hline
  \end{tabular}
  \caption{Ablations for TSEM on the PHOENIX14 dataset.} 
  \label{tab4} 
  \end{table}

\begin{table}[t]   
  \centering
  \begin{tabular}{ccc}
  \hline
  Configurations & Dev(\%) & Test(\%)\\
  \hline
  - & 21.2 & 22.3\\
  SSEM  & 19.9  & 21.4\\
  TSEM & 20.5 & 21.7 \\
  SSEM + TSEM & 19.8  & 21.4 \\
  TSEM + SSEM & 19.6  & 21.2 \\
  Parallelled & \textbf{19.5}  & \textbf{21.0} \\
  \hline
  \end{tabular}
  \caption{Ablations for the effectiveness of SSEM and TSEM on the PHOENIX14 dataset.} 
  \label{tab5} 
  \end{table}

\begin{table}[t]   
  \centering
  \setlength\tabcolsep{3pt}
  \begin{tabular}{lcc}
  \hline
  Methods & Dev(\%) & Test(\%)\\
  \hline
  - & 21.2 & 22.3\\
  w/ SENet~\cite{hu2018squeeze}  & 20.7  & 21.6 \\
  w/ CBAM~\cite{woo2018cbam} & 20.5 & 21.3 \\
  \hline
  CNN+HMM+LSTM~\cite{koller2019weakly} & 26.0  & 26.0 \\
  STMC~\cite{zhou2020spatial} & 21.1  & 20.7 \\
  C$^2$SLR~\cite{zuo2022c2slr} & 20.5  & 20.4 \\
  \hline
  SEN & \textbf{19.5} & \textbf{21.0} \\
  \hline
  \end{tabular}
  \caption{Comparison with other methods of channel attention or hand and face features on the PHOENIX14 dataset.} 
  \label{tab6} 
  \end{table}

\begin{figure}[t]
  \centering
  \includegraphics[width=\linewidth]{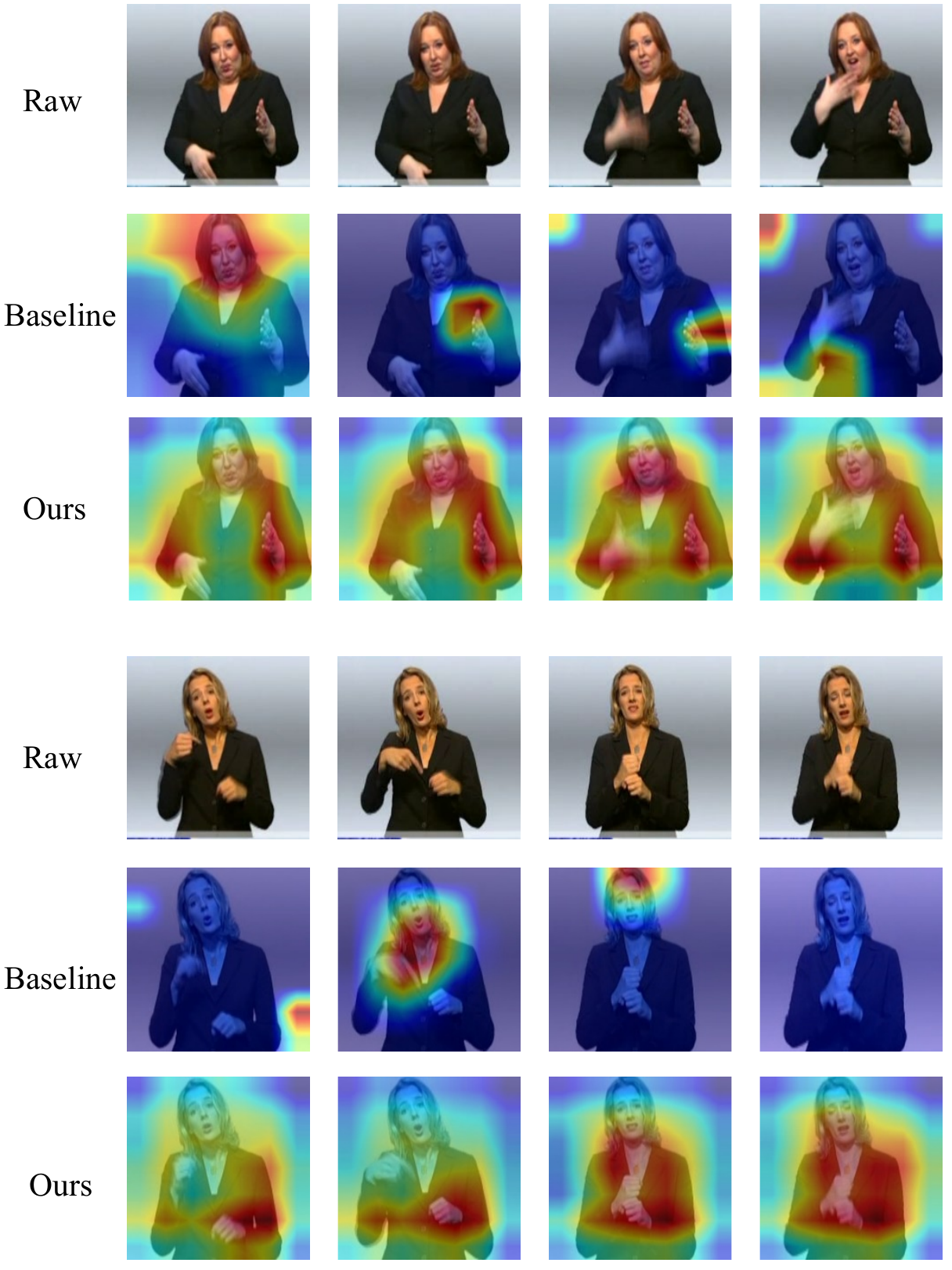}
  \caption{Visualizations of class activation maps by Grad-CAM~\cite{selvaraju2017grad}. Top: raw frames; Middle: class activation maps of our baseline; Bottom: class activation maps of our SEN. Our baseline usually focuses on nowhere or only attends to a single hand or face. Our SEN could generally focus on the human body (light yellow areas) and pays special attention to informative regions like hands and face (dark red areas).}
  \label{fig5}
  \end{figure}  

\subsection{Ablation Study}
We perform ablation studies on the PHOENIX14 dataset and report on both development (Dev) and testing (Test) sets. 

\subsubsection{Effects of the multi-scale architecture of SSEM.}
Tab.~\ref{tab1} ablates the implementations for the multi-scale architecture of SSEM. Our baseline achieves 21.2\% and 22.3\% WER on the Dev and Test Set. When fixing $K_t$=9, $K_s$=3 and varying the number of branches to expand spatial receptive fields, it's observed larger $N$ consistently brings better performance. When $N$ reaches 3, it brings no more performance gain. We set $N$ as 3 by default and test the effects of $K_t$. One can see that either increasing $K_t$ to 11 or decreasing $K_t$ to 7 achieves worse performance. We thus adopt $K_t$ as 9 by default. Notably, one can find SSEM brings few extra computations compared to our baseline. For example, the best-performing SSEM with $K_t$=9, $K_s$=3 and $N$=3 only owns 0.8M (\textless 0.1\%) extra FLOPs, which can be neglected compared to 364G FLOPs of our baseline model. Finally, we compare our proposed multi-scale architecture with a normal implementation of more computations. The receptive field of SSEM with $K_t$=9, $K_s$=3 and $N$=3 is identical to a normal convolution with $K_t$=9 and $K_s$=7. As shown in the bottom of tab.~\ref{tab1}, a normal convolution not only brings more computations than SSEM, but also performs worse, verifying the effectiveness of our architecture. 

\subsubsection{Implementations of SSEM to augment inputs features.} Tab.~\ref{tab3} ablates the implementations of SSEM to augment original features. It's first observed directly multiplying the attention maps $M_s$ with input features $s$ severely degrades performance, attributed to destroying input features distributions. Implemented in a residual way by adding $s$, $M_s \odot s + s$ could notably relieve such phenomenon and achieves +0.6\% \& +0.6\% on the Dev and Test Sets. Further, we first subtract $0.5\times \mathds{1}$ from the attention maps $M_s$ to emphasize or suppress certain positions, and then element-wisely multiply it with $s$. This implementation bring +1.0\% \& +0.8\% performance boost. Finally, we update this implementation in a residual way by adding input features $s$ as $(M_s -0.5\times \mathds{1}) \odot s +s $, achieving notable performance boost by +1.3\% \& +0.9\%.

\subsubsection{Study on TSEM.} Tab.~\ref{tab4} ablates the configurations for TSEM. We here adopt SSEM as our baseline and ablate the configurations for TSEM. It's first noticed that combining motion information by concatenating $u_r(t+1)-u_r$ with $u_r$ slightly outperforms only using $u_r$ to capture short-term temporal dependencies, verifying the effectiveness of local motion information. Next, when varying $P_t$, it's observed $P_t$=9 achieves the best performance among $P_t$=[7,9,11], which is adopted by default in the following.

\subsubsection{Study on the effectiveness of SSEM and TSEM.} Tab.~\ref{tab5} studies how to combine SSEM with TSEM. We first notice that only using SSEM or TSEM could already bring a notable performance boost, by +1.3\& +0.9\% and +0.7 \& +0.6\% on the Dev and Test Sets, respectively. When further combining SSEM with TSEM by sequentially placing SSEM before TSEM (SSEM+TSEM), placing TSEM before SSEM (TSEM+SSEM) or paralleling TSEM and TSEM, it's observed SSEM+TSEM performs best with +1.7\% \& +1.3\% performance boost on the Dev and Test Sets, respectively, adopted as the default setting.

\subsubsection{Comparison with other methods.} We compare our SEN with related well-known channel attention methods like SENet~\cite{hu2018squeeze} and CBAM~\cite{woo2018cbam}, and previous CSLR methods equipped with hand and face features by extra pose-estimation networks or pre-extracted heatmaps. In the upper part of tab.~\ref{tab6}, one can see SEN largely outperforms these channel attention methods, for its superior ability to emphasize informative hand and face features. In the bottom part of tab.~\ref{tab6}, it's observed SEN greatly surpasses previous CSLR methods equipped with hand and face features, even though they employ extra heavy networks or expensive supervision. These results verify the effectiveness of our SEN in leveraging hand and face features.

\subsection{Visualizations}
\subsubsection{Visualization for SSEM.}
We sample a few frames for expressing a gloss and plot the class activation maps for our baseline and SEN with Grad-CAM~\cite{selvaraju2017grad} in fig.~\ref{fig5}. The activation maps generated by our baseline usually focus on nowhere or only attend to a single hand or face, failing to fully focus on the informative regions (e.g., hands and face). Instead, our SEN could generally focus on the human body (light yellow areas), and pays special attention to those discriminative regions like hands and face (dark red areas). These visualizations show that without additional expensive supervision, our SEN could still effectively leverage the informative spatial features in a self-supervised way.

\begin{table*}[t]   
  \centering
  \setlength\tabcolsep{3pt}
  \begin{tabular}{ccccccccc}
  \hline
  \multirow{3}{*}{Methods} &\multirow{3}{*}{Backbone} & \multicolumn{4}{c}{PHOENIX14} & \multicolumn{2}{c}{PHOENIX14-T} \\
  & &\multicolumn{2}{c}{Dev(\%)} & \multicolumn{2}{c}{Test(\%)} &  \multirow{2}{*}{Dev(\%)} & \multirow{2}{*}{Test(\%)}\\
  & &del/ins & WER & del/ins& WER & & \\
  \hline
  Align-iOpt~\cite{pu2019iterative}& 3D-ResNet &12.6/2 & 37.1& 13.0/2.5 & 36.7 & -&-\\
  Re-Sign~\cite{koller2017re}& GoogLeNet&- & 27.1 &- &26.8 &- &-\\
  SFL~\cite{niu2020stochastic}& ResNet18 & 7.9/6.5 & 26.2 & 7.5/6.3& 26.8 & 25.1&26.1\\
  FCN~\cite{cheng2020fully}& Custom & - & 23.7 & -& 23.9 & 23.3& 25.1\\
  CMA~\cite{pu2020boosting} & GoogLeNet & 7.3/2.7 & 21.3 & 7.3/2.4 & 21.9  & -&-\\
  VAC~\cite{Min_2021_ICCV}& ResNet18 & 7.9/2.5 & 21.2 &8.4/2.6 & 22.3 &- &-\\
  SMKD~\cite{hao2021self}& ResNet18 &6.8/2.5 &20.8 &6.3/2.3 & 21.0 & 20.8 & 22.4\\
  \hline
  SLT$^*$~\cite{camgoz2018neural}& GoogLeNet  & - & - & - & - & 24.5 & 24.6\\
  CNN+LSTM+HMM$^*$~\cite{koller2019weakly}& GoogLeNet  & - &26.0 & - & 26.0 & 22.1 & 24.1 \\
  DNF$^*$~\cite{cui2019deep}& GoogLeNet  & 7.3/3.3 &23.1& 6.7/3.3 & 22.9 & - & -\\
  STMC$^*$~\cite{zhou2020spatial}& VGG11 & 7.7/3.4 &21.1 & 7.4/2.6 & 20.7 & 19.6 & 21.0\\
  C$^2$SLR$^*$~\cite{zuo2022c2slr} & ResNet18 & - & 20.5 &- & 20.4 & 20.2 & 20.4  \\
  \hline
  Baseline & ResNet18 & 7.9/2.5 & 21.2 &8.4/2.6 & 22.3 & 21.1 & 22.8\\
  \textbf{SEN (Ours)} & ResNet18 & 5.8/2.6  &\textbf{19.5} &   7.3/4.0 & \textbf{21.0}  & \textbf{19.3} & \textbf{20.7} \\
  \hline   
  \end{tabular}  
  \caption{Comparison with state-of-the-art methods on the PHOENIX14 and PHOENIX14-T datasets. $*$ indicates extra clues such as face or hand features are included by additional networks or pre-extracted heatmaps.} 
  \label{tab7}
\end{table*}

\subsubsection{Visualization for TSEM.} 
\begin{figure}[t]
  \centering
  \includegraphics[width=\linewidth]{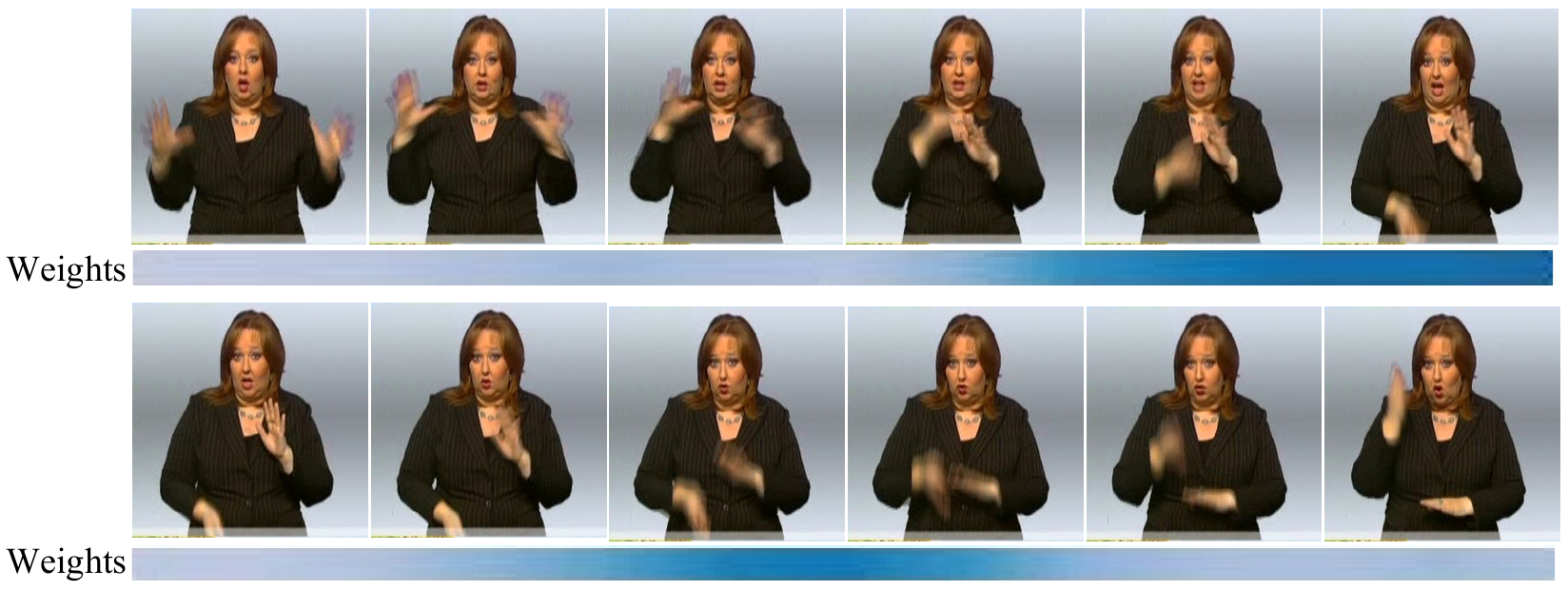}
  \caption{Visualizations of temporal attention maps for TSEM. One can find that TSEM highlight frames with rapid movements and suppress those static frames.}
  \label{fig6}
  \end{figure}

We visualize the temporal attention maps of TSEM in fig~\ref{fig6}. We sample several frames corresponding to an output gloss 'nord' as an example. The darker color, the higher weight. One can find that TSEM tends to allocate higher weights for frames with rapid movements (the latter two frames in the first line; the middle three frames in the second line). TSEM assigns lower weights for static frames with few body movements. Such observation is consistent with our habits, as humans always pay more attention to those moving objects in the visual field to capture key movements. Those frames can also be considered conveying more important pattern for expressing a sign.

\subsection{Comparison with State-of-the-Art Methods}
\textbf{PHOENIX14} and \textbf{PHOENIX14-T}. Tab.~\ref{tab7} shows a comprehensive comparison between our SEN and other state-of-the-art methods. We notice that with few extra computations, SEN could outperform other state-of-the-art methods upon both datasets. Especially, SEN outperforms previous CSLR methods equipped with hand and faces acquired by heavy pose-estimation networks or pre-extracted heatmaps (notated with *), without additional expensive supervision. 

\textbf{CSL-Daily}. CSL-Daily is a recently released large-scale dataset with the largest vocabulary size (2k) among commonly-used CSLR datasets, covering daily contents. Tab.~\ref{tab8} shows that our SEN achieves new state-of-the-art accuracy upon this challenging dataset with large progresses, which generalizes well upon real-world scenarios.

\textbf{CSL}. As shown in tab.~\ref{tab9}, our SEN could achieve extreme superior accuracy (0.8\% WER) upon this well-examined dataset, outperforming existing CSLR methods.
 
\begin{table}[t]   
  \centering
  \setlength\tabcolsep{2pt}
  \begin{tabular}{cccc}
  \hline
  Methods&  Dev(\%) & Test(\%)\\
  \hline
  LS-HAN~\cite{huang2018video}  & 39.0  & 39.4\\
  TIN-Iterative~\cite{cui2019deep}  & 32.8  & 32.4\\
  Joint-SLRT~\cite{camgoz2020sign}  & 33.1  & 32.0 \\
  FCN~\cite{cheng2020fully} & 33.2  & 32.5 \\
  BN-TIN~\cite{zhou2021improving} & 33.6  & 33.1 \\
  \hline
  Baseline & 32.8 & 32.3\\
  \textbf{SEN(Ours)} & \textbf{31.1} & \textbf{30.7} \\
  \hline
  \end{tabular}  
  \caption{Comparison with state-of-the-art methods on the CSL-Daily dataset~\cite{zhou2021improving}.} 
  \label{tab8}
  \end{table}

\begin{table}[t]   
  \centering
  \setlength\tabcolsep{2pt}
  \begin{tabular}{cc}
    \hline
    Methods&  WER(\%)\\
    \hline
    SubUNet~\cite{cihan2017subunets}   & 11.0\\
    SF-Net~\cite{yang2019sf} & 3.8 \\
    FCN~\cite{cheng2020fully}   & 3.0 \\
    STMC~\cite{zhou2020spatial}  & 2.1 \\
    VAC~\cite{Min_2021_ICCV} & 1.6 \\
    C$^2$SLR~\cite{zuo2022c2slr} & 0.9 \\
    \hline
    Baseline  & 3.5\\
    \textbf{SEN(Ours)} & \textbf{0.8} \\
    \hline
    \end{tabular}  
    \caption{Comparison with state-of-the-art methods on the CSL dataset~\cite{huang2018video}.} 
    \label{tab9}

  \end{table}

\section{Conclusion}
This paper proposes a self-motivated architecture, coined as SEN, to adaptively emphasize informative spatial and temporal features. Without extra expensive supervision, SEN outperforms existing CSLR methods upon four CSLR datasets. Visualizations confirm the effectiveness of SEN in leveraging discriminative hand and face features.
\bibliography{ref}
\end{document}